\documentclass[sigconf]{acmart}
\AtBeginDocument{%
  \providecommand\BibTeX{{%
    \normalfont B\kern-0.5em{\scshape i\kern-0.25em b}\kern-0.8em\TeX}}}

\copyrightyear{2021}

\acmYear{2021}
\setcopyright{rightsretained}
\acmConference[RecSys '21]{Fifteenth ACM Conference on Recommender Systems}{September 27-October 1, 2021}{Amsterdam, Netherlands}
\acmBooktitle{Fifteenth ACM Conference on Recommender Systems (RecSys '21), September 27-October 1, 2021, Amsterdam, Netherlands}\acmDOI{10.1145/3460231.3474264}
\acmISBN{978-1-4503-8458-2/21/09}

\usepackage{subfig}

\begin{document}

\title{Large-Scale Modeling of Mobile User Click Behaviors Using Deep Learning}

\author{Xin Zhou}
\affiliation{%
  \institution{Google Research}
  \city{Mountain View}
  \state{CA}
  \country{USA}}
\email{zhouxin@google.com}

\author{Yang Li}
\affiliation{%
  \institution{Google Research}
  \city{Mountain View}
  \state{CA}
  \country{USA}}
\email{liyang@google.com}


\begin{abstract}
Modeling tap or click sequences of users on a mobile device can improve our understandings of interaction behavior and offers opportunities for UI optimization by recommending next element the user might want to click on. %Accurate prediction of next UI element that the user wants to tap is an attractive yet challenging problem, which involves a multitude of factors such as the user's click history and current context, and often spans a large array of mobile apps. 
We analyzed a large-scale dataset of over 20 million clicks from more than 4,000 mobile users who opted in. We then designed a deep learning model that predicts the next element that the user clicks given the user's click history, the structural information of the UI screen, and the current context such as the time of the day. We thoroughly investigated the deep model by comparing it with a set of baseline methods based on the dataset. The experiments show that our model achieves 48\% and 71\% accuracy (top-1 and top-3) for predicting next clicks based on a held-out dataset of test users, which significantly outperformed all the baseline methods with a large margin. We discussed a few scenarios for integrating the model in mobile interaction and how users can potentially benefit from the model. %Our model also promises 66\% of decrease of user input needed for iterative traversal of a mobile UI in accessibility scenarios. 
\end{abstract}




\maketitle

\section{Introduction}

\label{sec:intro}

%\begin{figure}[t!]
%\includegraphics[width=1.0\textwidth]{figures/intro.png}
%\caption{Existing accessibility services allow a mobile user with dexterity impairments to "click" on a target by iterating through the actionable elements on the UI one by one via an external device by pressing on the Yes or No button. This can be time consuming to reach a target element because there are often many elements to traverse on the screen.

%\label{fig:intro}
%\end{figure} 

% Remove the Switch Access example with a Chrome example ???

%\begin{figure}[t!]
%\subfloat{\includegraphics[width=0.25\textwidth]{figures/chrome_label.png}}
%\label{fig:chrome}
%\caption{An example of click shortcut: Evaluate the probabilities of the actionable elements on screen to be click next and provide a shortcut the top-1 element.}
%\end{figure} 

User interaction behaviors or flows on modern mobile devices mainly consist of a sequence of taps (or clicks\footnote{We use "tap" and "click" interchangeably in this paper.}) on UI elements presented on the touchscreen, which result in a sequence of screen state transitions in response to these clicks. Modeling user click sequences is important for understanding rich mobile interaction behaviors, e.g., how screens transition and factors that are involved in these transitions \cite{10.1145/2037373.2037383}. %Particularly, similar to a language model that predicts next word in a natural language sentence, 
A click-sequence model that predicts which element the user would click next on the touchscreen can create opportunities for improving user experiences in many ways. For example, a mobile system can pre-fetch resources needed based on user action forecasts to reduce response latency (e.g., \cite{10.1145/2493432.2493490}) or enable adaptive user interfaces (e.g., \cite{DBLP:journals/ijmms/TanvirBCI11,10.1145/1518701.1518956}) by optimizing interfaces to facilitate next user input. In this paper, we focus on the computational modeling of click behaviors that can underpin these domains.

While previous work extensively investigated app-level prediction \cite{10.1145/2493432.2493490, 10.1145/1925019.1925023,10.1145/2642918.2647355, 10.1145/2370216.2370243, 10.1145/2037373.2037383, 10.1145/3180374.3181345, huangfu2015context, 8999259},
it is relatively sparse on finer-granularity user action prediction. In this paper, we focus on predicting next UI element the user clicks on. Prior work has attempted to address similar problems in various domains, e.g., mobile interaction \cite{8470114}, web surfing \cite{10.1145/3209978.3210060}, advertisement \cite{liu2015convolutional}, news \cite{wu-etal-2019-neural-news}, and shopping \cite{chen2019behavior, 10.1145/3383313.3412258, 10.1145/3159652.3159656}. However, previous approaches are inadequate in leveraging rich features that exist in mobile user click-through behaviors and addressing complex interactions between factors that are involved in interaction, which we intend to address in this paper.

%One important interaction scenario that motivates our work is mobile UI accessibility (see Figure \ref{fig:intro}). To support users with dexterity impairments who cannot direct-manipulate the touchscreen, an accessibility service such as Switch Access on Android \footnote{\url{https://support.google.com/accessibility/android/answer/6122836}} allows the user to iterate through all the actionable UI elements on the screen one by one until acquiring the target element. During the process, the user only needs to indicate whether the current element (highlighted with a bold bounding box) is the target item by pressing on the \texttt{Yes} or \texttt{No} button on a dedicated external hardware (see Figure \ref{fig:intro}). Clicking on the \texttt{Yes} button will select the current element and trigger a system action and the \texttt{No} button will move onto and highlight the next UI element on the screen. Currently, the actionable elements on the screen are sorted in a deterministic, spatial order (e.g., top-down). It can be time consuming to reach a target element especially when the target is at the end of the list---9 clicks are needed on average based on our analysis of a large-scale dataset. We hypothesize that predicting next element that users possibly use based on their behavior, instead of using a heuristic-based predetermined order, can effectively reduce the number of iterations needed of the user to reach a target element thus improves accessibility. 

There are two major challenges for accurate and scalable click sequence modeling. Firstly, there are a vast number of unique mobile apps today, which come with diverse UI screens and elements. Previous approaches rely on a predefined set of UI elements \cite{8470114} (similar to word tokens for language models \cite{10.5555/944919.944966}) so as to use conventional sequence modeling techniques such as LSTM \cite{10.1162/neco.1997.9.8.1735}. This kind of approach is difficult to scale to address diverse, ever growing UI elements out there. Secondly, user click behaviors can be drastically different, depending on many factors such as each user's usage history and situational factors encountered. Users often switch between apps \cite{10.1145/2037373.2037383} to carry out a task, which results in sequences spanning multiple apps. 

%On the other hand, though the large amount of UI elements, the possible next click is constrained in user's current screen. This motivates us to find a dedicated way to address this prediction problem.

In this paper, we present a novel approach, based on the advance of deep learning techniques \cite{seq2act, 10.5555/3295222.3295349, NIPS2015_5866}, for modeling user click sequences at scale. Our approach eliminates the need of using a predefined vocabulary of UI elements and provides an extensible architecture to incorporate a rich set of features, including the user's previous clicks and screens, the structural information of the current screen, and the current time as well as temporal dynamics between click events. We analyzed a large scale dataset of user click behaviors to gain insights for model design. The dataset involves over 20 million clicks from more than 4000 mobile users while using over 13,000 apps in their daily activity. 
%The click sequence of each user is flattened regardless of the application session boundaries. 
%In training, for recent historical screens, we encode the clicked UI elements with attention of other elements, and leverage a Transfer-based architecture to predict the next click on current screen. 
A thorough experiment based on the dataset shows that our model achieves 48\% and 71\% for top-1 and top-3 accuracy, which significantly outperformed the previous methods and also shows that click prediction is a challenging task. Our approach scales well as it does not involve complex feature engineering. We discuss interaction scenarios where our model can help and design considerations for integrating the model in an interaction flow. We share our insights into the model behavior and limitations, and plans for future work. The paper makes the following contributions.
\begin{itemize}
    \item An analysis of a large-scale dataset of mobile user click sequences that reveals rich factors and complexity in modeling click behaviors, which contributes new knowledge to understand mobile interaction behaviors.
    \item A Transformer-based deep model that predicts next element to click based on the user click history and the current screen and time. The model does not rely on a vocabulary of predefined UI elements and provides a general solution for modeling arbitrary UI elements for click prediction.
    \item A thorough experiment that compares our deep model with multiple alternative designs and baseline models, and an analysis of model behaviors and benefits that the model can bring to improve mobile interaction.
\end{itemize}

\section{Related Work}
% - Mobile user behavior analysis:
% - sleep with angry bird and many papers that cited this paper
% - next click prediction papers
% - predictive user interface
% - Reflection and other papers
% - a survey of deep learning approaches: LSTM and Transformer

Previous work has investigated mobile user behaviors based on interaction logs. B\"{o}hmer et al. discovered that multiple factors can impact the patterns of mobile app usage \cite{10.1145/2037373.2037383}. For example, the category of a mobile app and the user context such as the current time and location can influence the usage of mobile applications. Previous work also pointed out the average session with an app lasts less than a minute, even through users may spend a long time using their phones. These findings are aligned with our analyses based on a large-scale dataset.

Previous work has extensively investigated approaches for modeling app usage and predicting next app that are likely to be used, e.g., \cite{10.1145/2684822.2685302, 10.1145/2370216.2370243, 10.1145/2493432.2493490}. Particularly, prior work \cite{10.1145/2370216.2370442, 10.1145/3180374.3181345} revealed that time signals such as the hour of the day and the day of the week as well as recent usage history are essential for predicting next app usage. Choonsung et al. \cite{10.1145/2370216.2370243} developed various ways for integrating prediction models into the mobile phone homescreen that greatly reduced search time for apps. Li proposed app prediction as a general service on mobile devices \cite{10.1145/2642918.2647355}. App prediction has become available on a variety of mobile platforms that is frequently in use by mobile users for daily app launches.

Compared to app prediction, there is relatively less prior work on finer-granularity user action prediction especially for mobile interaction. Deep learning has been broadly used in various domains, which has shown promise in addressing complex problems without extensive feature engineering. For interaction behavior modeling, sequence models such as LSTMs (e.g., \cite{8470114, 10.1145/3173574.3173603}) have been extensively used because of their extensive architecture and the ability to capture long-range dependencies. Convolutional neural networks have also been used for modeling sequences of multiple recommendation tasks \cite{liu2015convolutional, 10.1145/3159652.3159656}. Transformer \cite{10.5555/3295222.3295349} is a more recent model for sequence modeling that has produced state-of-art results on various tasks. We next focus on recent work that are particularly relevant to our work.

Lee et al. developed a model to predict the element that the user is likely to click on the current screen based on the user's previous click behaviors \cite{8470114}. To use conventional sequence modeling techniques such as LSTMs \cite{10.1162/neco.1997.9.8.1735}, the prior work
%pre-processed The researchers pre-process the clicks sequence data by splitting by application sessions, and then models the click prediction problem with LSTM, in a similar way as language modeling. 
treated user click behaviors as a sequence of tokens where each token is derived from a UI element, which requires a predefined vocabulary of elements. While an important contribution of the prior work was the mechanism for identifying an element to define the vocabulary, this approach is fundamentally limited because mobile apps are fast growing and their UIs are quickly evolving. It is impractical to identify every element in an ever growing collection of UIs. In addition, the previous work only models click sequences within each app and does not handle cross-app transitions where the latter constitute 26\% of click behavior based our data analysis. In our work, we deliberately address the representation challenge of UI elements. Without requiring a predefined vocabulary of elements, rather we use neural pointers and contextual representations of elements. Our model is also designed to easily leverage a richer set of features than treating each click as a token, and naturally handle both within and cross-app transitions.

Transformer is employed in previous works like Chen et al. for shopping recommendation \cite{chen2019behavior}, and Wu et al. in news recommendation \cite{wu-etal-2019-neural-news}. Their model takes input of embeddings of previously clicked items and a candidate next item, and outputs a score to rank each candidate. Our model, which is also based on Transformer, has several critical differences. First, our model uses a different set of features for click prediction across arbitrary apps, instead of a single app that the previous work focused on. Second, we use a hierarchical Transformer to model both in the context of a screen and a click sequence. Third, we use neural pointer to find next item instead of using a ranking model (i.e., sigmoid output in \cite{chen2019behavior} or inner product in \cite{wu-etal-2019-neural-news}). We extensively experimented with our model in comparison with  alternative methods including these deep models. Lastly, in contrast to previous work (e.g., \cite{DBLP:journals/ijmms/TanvirBCI11,10.1145/1518701.1518956}) on human factors and design options for adaptive UIs, we concentrate on computational modeling of user click behaviors, which provides a predictive model basis for adaptive UIs.

\section{Understanding Mobile Click Behaviors}
% - User Click Behaviors
% - features in the data
To develop computational models for predicting next element to tap, we first analyze a large-scale dataset of mobile click behaviors and gain insights into these behaviors. The analyses of the dataset inform the design of our deep model.

%Mobile users interact with the system mainly by clicks on screen of devices. The context of a click action includes current screen, local time and which application using (unfortunately geographic information is unavailable for our dataset). The current screen compose of all the information visible on the screen of the mobile device. It will change when a system internal event or user action occurs. Our work focus on user actions and the screen information shown to user right before users make actions. Time and package name of applications are also part of current context. Some researchers \cite{7869347} point out that these context info influence user behavior in a variety of ways, and our experiments also show that they are essential for the prediction of our model, see \ref{sec:analysis}.

\subsection{Data Processing}

User volunteers were sampled from a consented research panel to be representative of the mobile user population and opted in to participate in multi-week study. The interaction events were collected during their regular usage of device through additional software that captured data from app UIs.

%User volunteers opted in for a multi-week data collection, and they were sampled to be representative of the mobile user population. Volunteers needed to explicitly consent to a privacy policy and the data being collected. The interaction events were anonymously collected during their regular usage of device through Android Accessibility APIs that record both user actions and screen information. \footnote{\url{https://developer.android.com/reference/android/view/accessibility/package-summary}}. 
Each event consists of the user input (e.g., a \texttt{Click} versus a \texttt{Swipe}), the UI element that the input is applied to as well as a structural representation of the entire screen where the event takes place (i.e., the View Hierarchy\footnote{\url{https://developer.android.com/reference/android/view/View.html}}). %We do not collect any other customized gesture events besides taps.

A UI element is considered \textit{actionable} only if its following attributes in the view hierarchy are set to true: \texttt{clickable}, \texttt{visible}, and \texttt{enable}. For each UI element, we extract its text content, type, and bounding box positions on the screen from the view hierarchy. These properties will be used for featurizing the element later for modeling. We extract the text content of an element by looking at its \texttt{text} property first. If it is absent or empty, we use its \texttt{content-desc} attribute. The last resort is to extract text content from the \texttt{resource-id} property of the element.

The click events of each user are sorted  chronologically as a sequence. Each event consists of the screen view hierarchy that the click occurs, the element on the screen that is being clicked and the time of the event. %While user behavior can be location-dependent, location information is absent in our dataset. 
We filtered sequences of user participants who had outlier behaviors, such as sporadic usage of their devices or a large number of clicks produced in a short period of time. As a result, the dataset consists of over 20 million clicks, which form click sequences from more than 4,000 Android users using over 13,000 unique apps on their smartphones. 

\subsection{Data Analysis}

%\begin{figure}
%\subfloat{\includegraphics[width=.5\columnwidth]{figures/seq_timespan.png}} 
%\subfloat{\includegraphics[width=.5\columnwidth]{figures/seq_length.png}}
%\caption{The time span and the sequence length distribution for each click sequence in our dataset.}
%\label{fig:seqence_distirbution}
%\end{figure}

%\begin{figure}
%    \centering
%    \includegraphics[width=0.8\textwidth]{figures/clicks_by_week.png}
%    \caption{The distribution of sequences (users) based on the average number of clicks per week for each sequence.}
%    \label{fig:click_by_week}
%\end{figure}

\subsubsection{Sequence Lengths \& Time Spans} 
The length of these sequences varies from 100 to 25,000 clicks with the time span of each sequence ranging from 2 to 6 weeks (Mean=4.12 and Std=0.62). % (see Figure \ref{fig:seqence_distirbution}), as we use data collected around Aug of 2019. 
One factor that contributed to the variance in click numbers is the difference between power users versus novice users, which ranges from 19 clicks to over 5000 clicks per week (Median 833). % (see Figure \ref{fig:click_by_week}). 
These results showed that the dataset covers a wide range of user click behaviors that provide a solid basis for developing and evaluating machine learning models.

\subsubsection{UI Complexity}
\begin{figure*}
{\includegraphics[width=0.8\textwidth]{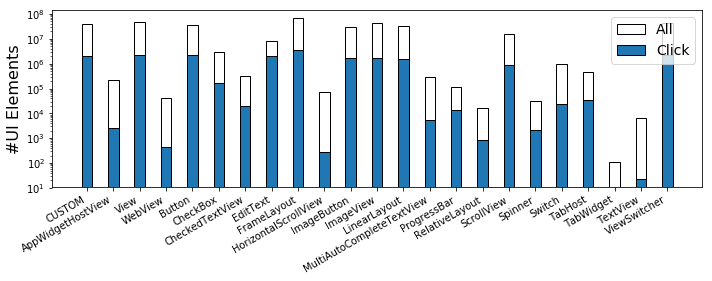}} 
\caption{The distribution of the total number of elements versus the clicked ones for each UI type, based on the dataset. The count of UI elements (the Y axis) is displayed at the $Log$ scale.}
\label{fig:obj_type}
\end{figure*}

The dataset involves a diverse set of UI elements with 24 types (see Figure \ref{fig:obj_type}). Overall, for all the actionable UI elements that we analyzed, the number of clicks that each type of element received is well aligned with the total number of elements of each type that exist in the UI screens in the entire dataset, with a few exceptions. For example, \texttt{TextView} received relatively few clicks and \texttt{TabWidget} received none, which is based on the usage of the set of user participants during the period of data collection.

\begin{figure*}
\begin{minipage}{0.33\textwidth}
  \centering\captionsetup{width=0.8\linewidth}
  \includegraphics[width=\textwidth]{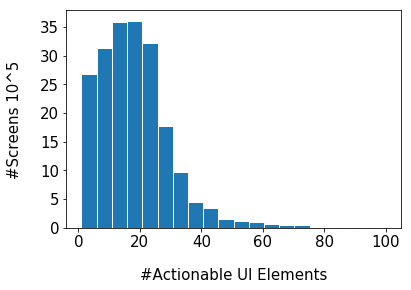}
  \caption{The number of actionable elements on each screen.}
  \label{fig:screen_distribution}
\end{minipage} %\hfill% or \hspace{5mm} or \hspace{0.3\textwidth}
% \hspace{0.5em}
\begin{minipage}{0.33\textwidth}
  \centering\captionsetup{width=0.8\linewidth}
  \includegraphics[width=\textwidth]{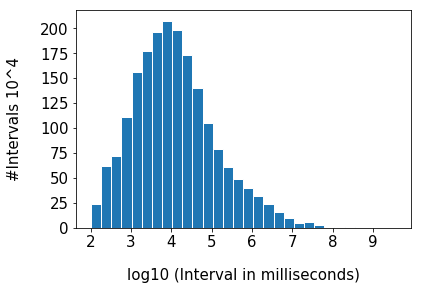}
  \caption{The distribution of time intervals between click events.}
  \label{fig:gaps}
\end{minipage}
\begin{minipage}{0.33\textwidth}
  \centering\captionsetup{width=1.0\linewidth}
  \includegraphics[width=\textwidth]{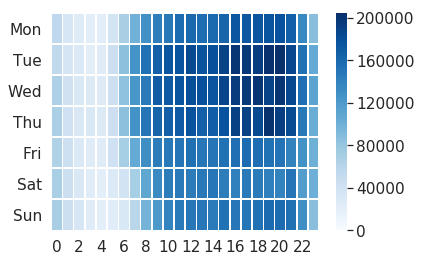}
  \caption{The distribution of click events based on the day of a week and the hour of a day when an event occurred.}
  \label{fig:click_time}
\end{minipage}
\end{figure*}

There are often a large number of actionable elements on each screen in this dataset: Mean=18, Std=12 (see Figure \ref{fig:screen_distribution}). The number of actionable elements on a screen determines how challenging it is for a model to predict the next element to be clicked by the user from these candidates. 

% \begin{figure}
%     \subfloat[The distribution of the number of actionable elements on each screen.]{\includegraphics[width=0.48\linewidth]{figures/screen_size.png}}
%     \label{fig:screen_distribution}
%     \hspace{0.5em}
%     \subfloat[The distribution of time intervals between click events.]{\includegraphics[width=0.48\linewidth]{figures/gaps.png}}
%     \label{fig:gaps}
% \end{figure}

\subsubsection{Temporal Patterns}
\begin{figure*}
\centering
\begin{minipage}{.53\textwidth}
  \centering\captionsetup{width=1.0\linewidth}
  \includegraphics[width=\columnwidth]{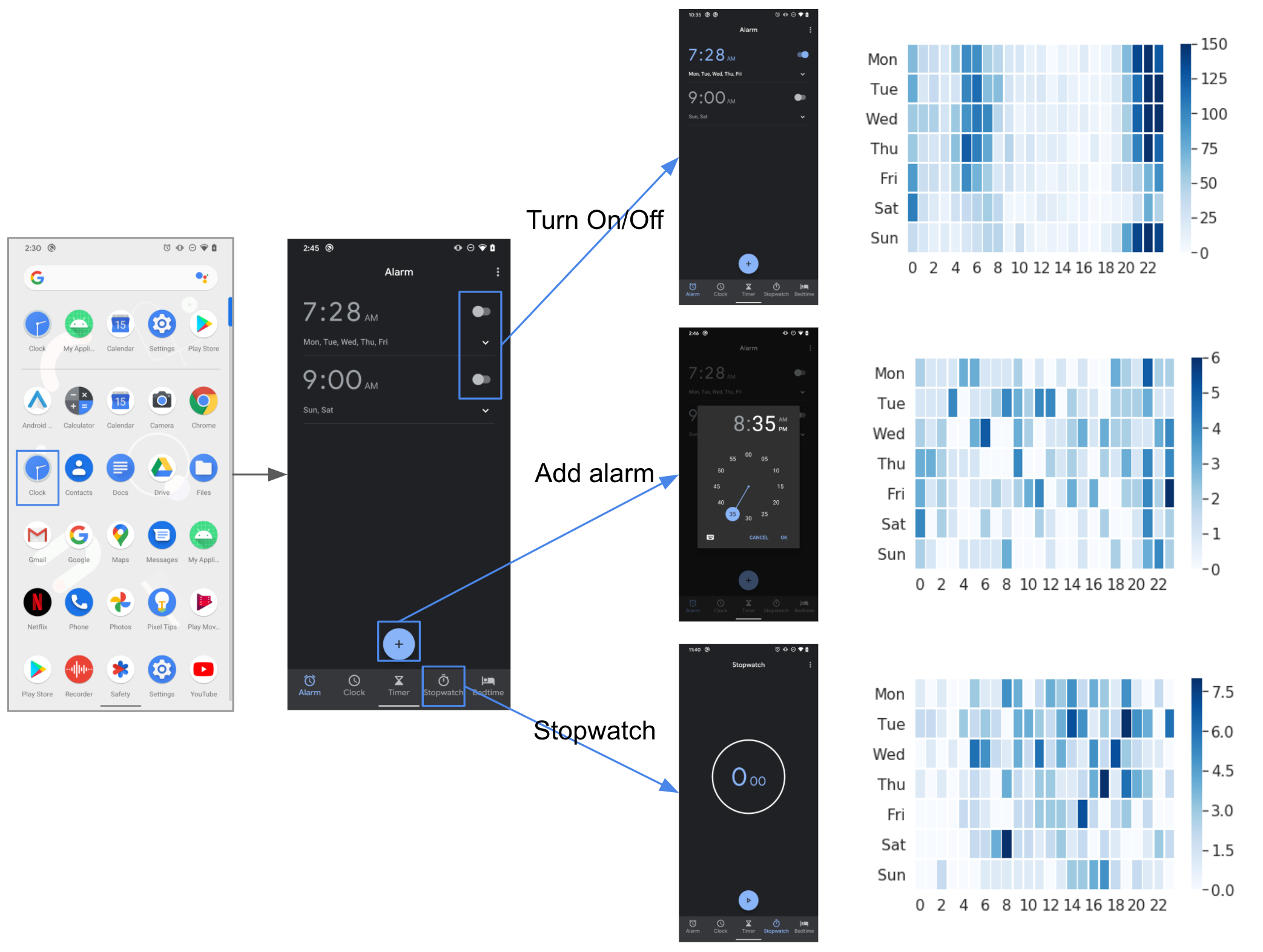}
  \caption{The click-transition sequences show that a user clicked on different elements of the Clock screen in a different hour and day. The toggle buttons of alarms are mostly used in the nights Sunday through Thursday, when the following day is a workday. But the stopwatch is used evenly through the week.}
  \label{fig:example_clock}
\end{minipage}
\begin{minipage}{.42\textwidth}
  \centering\captionsetup{width=0.9\linewidth}
  \includegraphics[width=\columnwidth]{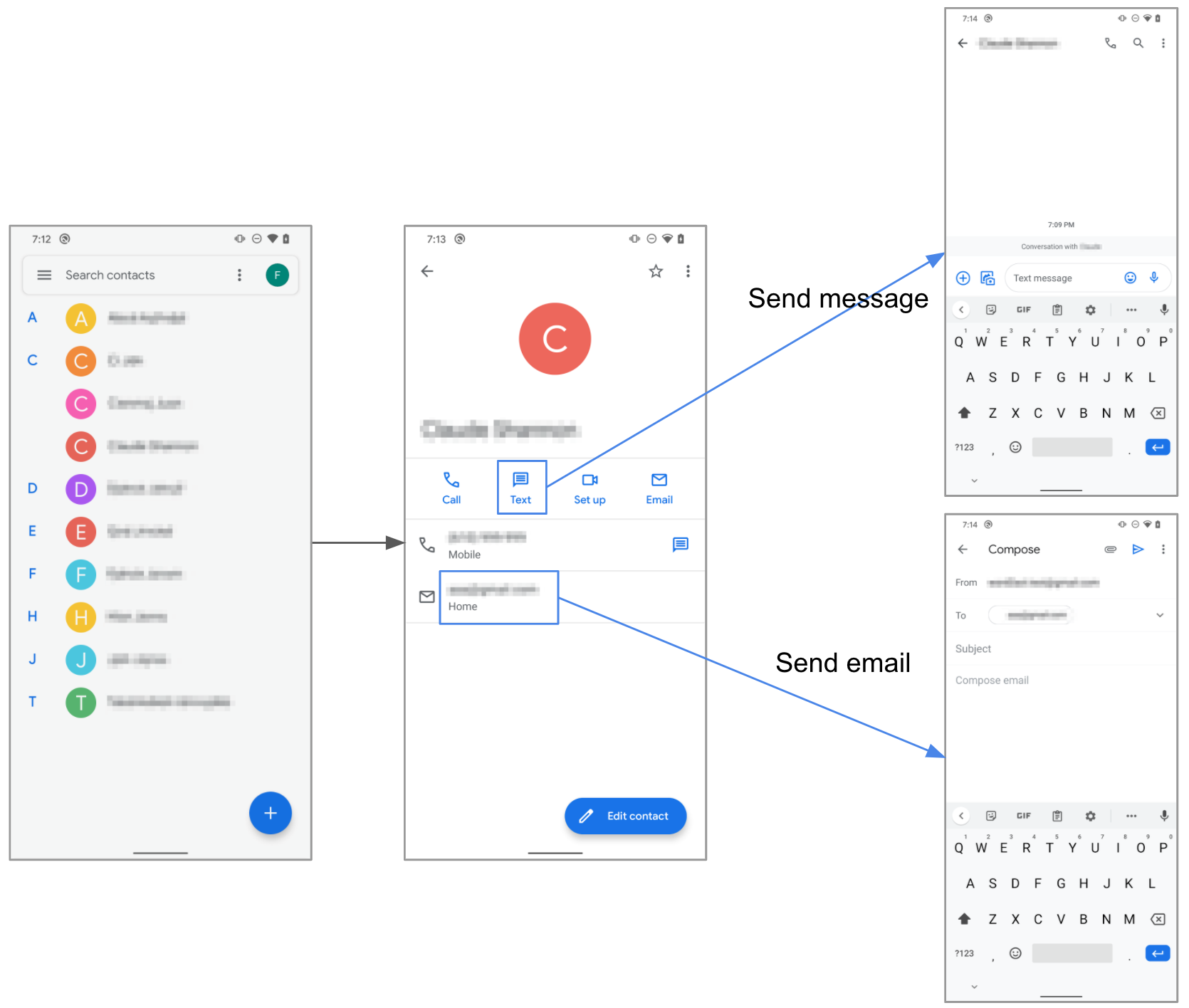}
  \caption{An example of cross-apps click transitions. The "text" and "email" elements in Contacts will lead the user to the Messaging app and the Gmail app respectively.}
  \label{fig:example_contacts}
\end{minipage}
\end{figure*}

The time analysis of click events involves two aspects: one is the temporal dynamics between events, and the other is the temporal regularity. For the temporal dynamics, we analyzed the time intervals between events (see Figure \ref{fig:gaps}). We can see most event transitions have a short time span. Intuitively, the shorter a time span is for an transition, the more relevant the two adjacent events are. For analyzing the temporal regularity, we computed the hour of the day and the day of the week when an event occurred using both the raw timestamp and the timezone information of an event. Figure \ref{fig:click_time} shows how events are distributed across hours of a day and days of a week. As expected, there are fewer events on the weekends than on the weekdays. Across hours of a day, there is a heavier usage of phones in the late afternoons and evenings. These results show that the dataset sufficiently covers users' daily usage of mobile devices. The example in Figure \ref{fig:example_clock} shows that click behaviors can be highly time-dependent. We intend to use both temporal signals in our deep model.

\subsubsection{Click Transitions}
A click event triggers the transition from one screen to another, which can be a popup menus overlaying the current screen or a completely new screen. About 30\% of these events were generated from popular apps such as Chrome, SMS and YouTube. However, there are a vast number of long-tail apps that play a vital role in users' daily mobile usage. While many of these transitions occur within an app, i.e., the current screen and the next screen after an click both belong to the same app, 26\% of transitions do occur across apps, (see Figure \ref{fig:example_contacts} for an example). This observation is aligned with previous findings \cite{10.1145/2037373.2037383} that users tends to transition across apps frequently. This implies that it is important for the model to be able to handle cross-app transitions.

% \begin{figure}
% \subfloat[By package name]{\includegraphics[width=.63\columnwidth]{figures/app_pie.png}} 
% \subfloat[By session length]{\includegraphics[width=.37\columnwidth]{figures/session_pie.png}}
% \caption{Distribution of the 20,362,211 screens/clicks by package name (a), or by the length of session they belong to. This shows the diversity and complexity of user click behavior: user clicks in a lot of applications, and switch among them frequently.}
% \label{fig:app_distribution}
% \end{figure}

%\begin{figure}
%    \centering
%    \includegraphics[width=1.0\columnwidth]{figures/app_pie.png}
%    \caption{Distribution of the 20 million screens by package name.}
%    \label{fig:app_distribution}
%\end{figure}

%\begin{figure}
%    \centering
%    \includegraphics[width=0.6\columnwidth]{figures/session_pie.png}
%    \caption{Distribution of the 20 million screens by the length of session they belong to.}
%    \label{fig:session_len}
%\end{figure}

\section{Modeling Click Behaviors}
\label{sec:model}

With the analytical understanding of mobile user click behaviors, we first formulate our modeling task, and then describe the design of our deep model. 

\subsection{Problem Formulation}
As stated in the previous section, a click event consists of a tuple of attributes, including the screen, the element being clicked on the screen, the time that the event occurs, and the app that the screen belongs to. The $i$th event in a click sequence is formulated as the following (see Equation \ref{eq:event}).
\begin{equation}
    \label{eq:event}
    e_{i}=[s_i,c_i,t_i,a_i]
\end{equation}

where $s_i$ is the screen that contains a collection of UI elements $o_{i,j}\in{s_{i}}$ where $1\leq{j}\leq{|s_{i}|}$ and $|s_{i}|$ represents the number of elements in the screen $s_{i}$. $c_i$ denotes the index position of the element being clicked on the screen, in the pre-order traversal of the view hierarchy tree (sibling elements are spatially ordered). Consequently, the element being clicked is represented as the following.
\begin{equation}
    o_{i,c_{i}}=\mbox{Preorder\_Traversal}(s_i)[c_i]
\end{equation}

$t_i$ represents the temporal information of the event that consists both temporal regularity, i.e., the hour of the day, $r_i$, and the day of the week, $w_i$, and temporal dynamics, i,e., how distant the event is with respect to the prediction time, denoted as $v_i$. $a_i$ identifies the app that the screen belongs to. A click sequence of a user is thus represented as a sequence of these tuples.
\begin{equation*}
    e_{1:n}=\{(s_1,c_1,t_1,a_1),(s_2,c_2,t_2,a_2),...,(s_n,c_n,t_n,a_n)\}
\end{equation*}
So the click prediction problem is defined as the following.
\begin{equation}
\label{eq:target_idx}
    c_{n+1}=f(s_{n+1},r_{n+1},w_{n+1},a_{n+1},e_{1:n})
\end{equation}
Given a sequence of previous events $e_{1:n}$ and the current context (i.e., the current screen $s_{n+1}$, the current hour and day $r_{n+1}$ and $w_{n+1}$, the current app $a_{n+1}$), we want to predict which element on the screen, $c_{n+1}$, is likely to be clicked by the user next, i.e., at the $(i+1)$th step. In this work, we intend to use a deep model to realize function $f$.

\begin{figure*}
\centering
  \includegraphics[width=1.0\textwidth]{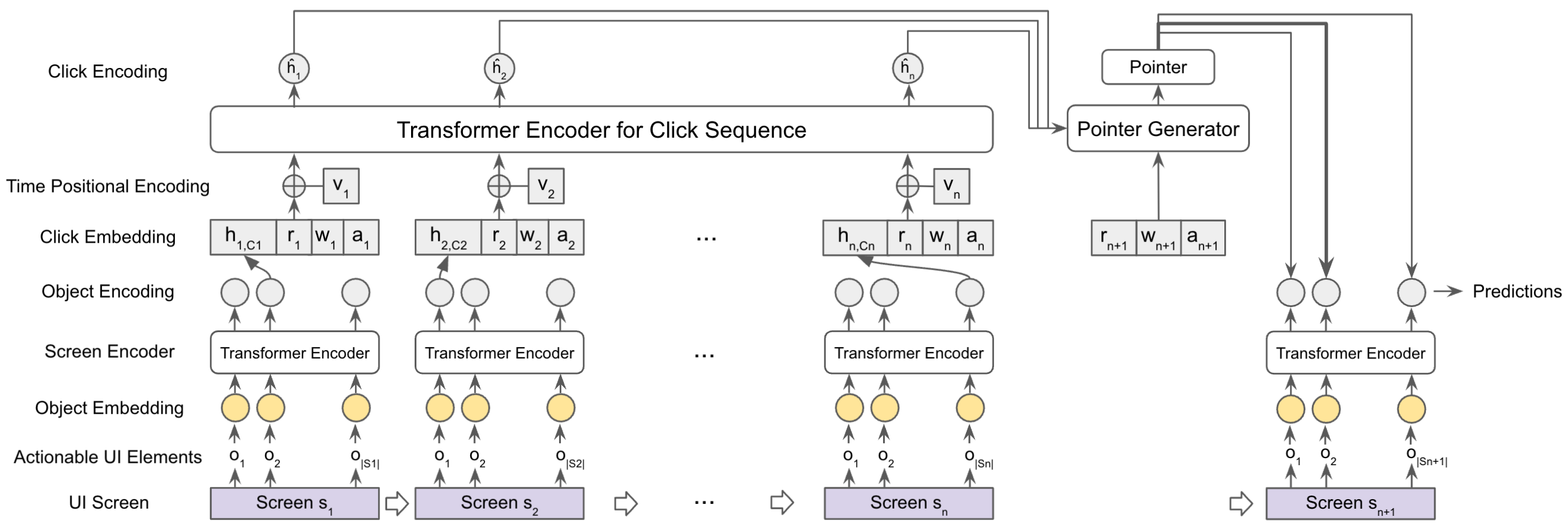}
  \caption{The architecture for our deep model for predicting the element to be clicked on a given screen. App IDs $a_i$ are optional in our model, and our model is app agnostic and provides a general solution for modeling clicks from arbitrary UIs.}
  \label{fig:model}
\end{figure*}

\subsection{The Design of the Deep Model}
There are three tasks for designing the prediction model. First, we need to derive a semantic representation of the clicked element of each event in the context of its screen. Second, we need to model a sequence of click events so as to encode the user's history. Last, based on the encoding of previous events and the current context, we want to design the model to select the element that is most likely to be clicked from all the actionable elements on the current screen. We design our deep model as hierarchical transformers coupled with neural pointers (see Figure \ref{fig:model}).

\subsubsection{Encoding a UI Element in the Context of its Screen}
We use Transformer \cite{10.5555/3295222.3295349}, a neural attention-based model that has led to state-of-the-art results for many problems. Specifically, we use the Transformer Encoder to encode each element in the screen in a similar way to previous work \cite{seq2act} (see Equation \ref{eq:encode_screen}). The self-attention in Transformer Encoder allows each element to be represented in the context of all the elements on the screen (see \cite{10.5555/3295222.3295349} for more details about self attention). 

\begin{equation}
\label{eq:encode_screen}
    h_{i,1:|s_{i}|}=\mbox{Transformer\_Encoder}(\mbox{Embed}^{o}(o_{i,1:|s_{i}|}),\theta)
\end{equation}

\noindent $\theta$ is the trainable parameters. For all the elements on screen $s_i$, we first compute the embedding of each element, $\mbox{Embed}(o_{i,1:|s_{i}|})$, based on its text content, the type attribute, and the bounding box (position and size) on the screen. The text content and the type attribute form the content embedding of an element while the bounding box positions form the positional embedding of the element.

Each word in the text content of an element is represented via a regular word embedding, and the text content of an object, which can contain more than one word, is represented as the average embedding of all the words, similar to a bag of words. The type is a categorical value and also represented as an embedding vector, which is combined with the text embedding to form the content embedding of the element.

%Object embedding is a combination of the \textit{content embedding} and \textit{positional encoding} of each UI object.

The positional embedding of an element is computed based on the bounding box positions of the element: \texttt{[left, top, right, bottom]}. We simply normalized each coordinate value to an integer in the range of $[0, 100)$, which is then represented as an trainable embedding vector. The sum of positional and content embeddings forms the input of the Transformer Encoder model.

The embeddings of all the elements on the screen are then fed to a Transformer Encoder (Equation \ref{eq:encode_screen}), which results in a latent fixed-dimensional representation for each element, $h_{i,j}\in{R^{d}}$ where $1\leq{j}\leq{|s_{i}|}$ as defined earlier, and $d$ is the depth of the latent vector. In particular, $h_{i,c_{i}}$ contextually represents the UI element being clicked on at step $i$. Such a representation of a clicked element is contextual because the self attention in Transformer Encoder learns to relate an element to all other elements on the screen based on both their content similarity and spatial adjacency.

\subsubsection{Encoding a Click Sequence}
With each clicked element semantically represented as $h_{i,c_{i}}$, we next describe how we model a sequence (history) of click events, $e_{1:n}$ (see Equation \ref{eq:event}). For this task, we again design our model using Transformer Encoder. 

As revealed in our data analysis, the temporal information, $t_i$, is an important factor for mobile click behaviors, and there are two aspects to it: temporal regularity and dynamics. For temporal regularity features, i.e., the hour of the day, $r_i$, and the day of the week, $w_i$, of an event, we represent them as a one-hot vector of size 7 (for 7 days of a week), and 24 (for 24 hours of a day) respectively, from which we acquire a trainable embedding vector for each feature. Similarly, we represent the app the screen belongs to, $a_i$, via an embedding vector as well. Along the clicked element representation, $h_{i,c_{i}}$, we form the embedding of a click event as the following (Equation \ref{eq:click_embed}).

\begin{equation}
\label{eq:click_embed}
    c_{i}^{E}=[h_{i,c_{i}}; \mbox{Embed}^{r}(r_{i}); \mbox{Embed}^{w}(w_{i});
    \mbox{Embed}^{a}(a_{i})]W_{c}
\end{equation}

We first concatenate all these embedding vectors with the clicked element encoding. The concatenation is then linearly projected via trainable parameters, $W_{c}$, to a dimension $d$, which results in the click event embedding, $c_{i}^{E}\in{R^{d}}$.

For temporal dynamics, specifically the time interval between an event in the past and the prediction time, $v_i$, we use the information to derive the positional encoding of each event in the history sequence. Notice that this differs from the positional encoding schema in the original Transformer case where events are evenly distributed along the time dimension. As shown in Figure \ref{fig:gaps}, the time interval between click events can vary drastically. The temporal positional encoding of an event should capture how recent the event is for the current step. We compute this elapsed time-based positional encoding as the following.
\begin{equation}
    v_{i}^{E}=\mbox{Embed}^{v}(\mbox{Floor}(log(v_{i})))
\end{equation}

where we first take the logarithm of $v_{i}$, the elapsed time from the $i$th step to the current $(n+1)$th step, and then discretize it using its floor value. This again is represented using a vector through the learnable embedding function $\mbox{Embed}^{v}$. Finally, we combine both the click event embedding (Equation \ref{eq:click_embed}) and its time positional encoding to form the input to the Transformer Encoder by adding them: $e_{i}^{E}=c_{i}^{E} + v_{i}^{E}$. We then feed the event embeddings to another Transformer Encoder to generate the latent representation of each click event, $\hat{h}_{i,1:|s_i|}$. $\beta$ is trainable parameters in the Transformer Encoder model.

\begin{equation}
    \label{eq:evt_latent}
    \hat{h}_{i,1:|s_i|}=\mbox{Transformer\_Encoder}(e_{i,1:|s_i|}^{E},\beta)
\end{equation}

%As show in Figure \ref{fig:model}, the input to Sequence Transformer Encoder, named \textit{click encoding} in graph, compose of three parts: 1) the clicked UI object encoding plus \textbf{sequence positional encoding} $pos_{1:n-1}$, 2) time encodings $t_{1:n-1}$ including day-of-week and hour-of-day, and 3) package name encoding $p_{1:n-1}$:

\subsubsection{Predicting Next Element via Neural Pointer}

With the click history represented, we now discuss how we can find the element that is most likely to be clicked on the current screen, $s_{n+1}$, given the current hour and day $r_{n+1}$ and $w_{n+1}$ as well as the app the screen belongs to, $a_{n+1}$. First of all, we compute a contextual representation of each element on the current screen, $h_{n+1,1:|s_{n+1}|}$, which is the same way as how we compute the representation of each element in the previous screens (see Equation \ref{eq:encode_screen}). We then compute a "pointer" to point into the current screen---that evaluates how likely each element would be clicked next. The "pointer" is realized via a $M$-layer perceptron that takes the current hour and day, $r_{n+1}$ and $w_{n+1}$, as well as the app, $a_{n+1}$, as the input, and at the same time attends to the latent representation of each event in the history, $\hat{h}_{i,1:|s_i|}$, via multi-head attention \cite{10.5555/3295222.3295349}.

\begin{equation}
    q_{n+1}^{0}=[\mbox{Embed}^{r}(r_{n+1});\mbox{Embed}^{w}(w_{n+1});\mbox{Embed}^{a}(a_{n+1})]
\end{equation}
\begin{equation}
\label{eq:pointer_generator}
    q_{n+1}^{m+1}=\mbox{MultiHead\_Attention}(q_{n+1}^{m},\hat{h}_{i,1:|s_i|}),\theta^{q})
\end{equation}

where $m$ denotes the layer index and $0\leq{m}\leq{M}$. The final output of the pointer generator is thus $q_{n+1}^{M}$. We then compute how well the pointer vector $q_{n+1}^{M}$ matches the latent representation vector of each element on the current screen using a typical neural alignment mechanism.

\begin{equation}
    \alpha_{n+1,j}=q_{n+1}^{M}W^{q}\cdot h_{n+1,j}
\end{equation}

where we first linearly project the pointer vector $q_{n+1}^{M}$ via $W^q$ (learnable parameters), and then perform dot product between the linear projection of the pointer and the latent representation vector of each element. The alignment score $\alpha_{n+1,j}$ is a scalar value, which can then be used to calculate the probability of each element to be clicked using a softmax.

\begin{equation}
\label{eq:softmax}
    \mbox{Prob}_j=\frac{\mbox{exp}(\alpha_{n+1,j})}{\sum_{1\leq{k}\leq{|s_{n+1}|}}\mbox{exp}(\alpha_{n+1,k})}
\end{equation}

With Equation \ref{eq:softmax}, we can find the element that is most likely to be clicked, i.e., $c_{n+1}=\mbox{argmax}_{j}Prob_j$, which realizes the Equation \ref{eq:target_idx}. The entire model can be trained end to end by minimizing the cross entropy loss between the probability distribution of next element to click (Equation \ref{eq:softmax}), and the element that is actually being clicked on the screen---the groundtruth.

Note that our model does not rely on app specific information, except the app identifier $a_i$. As we will discuss later, our experiments revealed that using the app identifier, $a_i$, does not significantly improve our accuracy. As a result, the design of our entire model can be app independent, which means that it can handle UIs from arbitrary, unseen apps.

\section{Experiments}
% - Data processing (or mention earlier)
% - Hyperparameters 
% - Experiment setups
% - Metrics
% - Report on results
To evaluate the accuracy of our model, we conducted a thorough experiment by comparing our model with multiple baselines based on a set of evaluation metrics. 
%In this section, we first d our dataset and discuss how we further process the dataset for training, validation and testing. We next discuss the set of metrics we focus on, which is then followed by a description of baseline models we experimented with. We then report on the accuracy of these models.

\subsection{Dataset}
We randomly split the dataset into 80\% of the sequences for training, 10\% for validation to tune hyper parameters of the models and 10\% for testing. %The word vocabulary size is 50K.
Several baseline methods that we compare with require each screen to be uniquely identified via a static ID. To this end, we compute a hash ID for each screen. To generate a hash ID for a screen, we concatenate the features of each UI element sequentially according to the preorder traversal of the screen's view hierarchy tree. We then map the concatenation to a hash signature using the MurmurHash3 algorithm\footnote{\url{https://guava.dev/releases/snapshot/api/docs/com/google/common/hash/Hashing.html\#murmur3_128--}}. 
%Screens with nonessential, minor differences, such as system clock time changes at the top right corner of the screen, are considered the same. 

%Coordinates of a UI element include [x1, y1, x2, y2] values, and are normalized to range [0, 100) by the coordinates of the root UI element in the top level of view hierarchy tree. These actionable UI elements are potential \textit{objects} for user clicks.

\subsection{Baselines}
\label{baselines}
We compare our model with 9 alternative methods in our experiment. 3 of these methods are heuristics-based, which offer a baseline for understanding the difficulty of the task. 3 other methods are based on classic ML methods that have been widely used in the % HCI field for
recommendation/prediction tasks. We also included 3 recent methods that used deep learning models for click prediction.

%\subsubsection{Spatial Sorting} 
%To understand how a An intuitive approach for iterating through UI elements on the screen is to follow a spatial order. Existing accessibility services such as Android Switch Access\footnote{\url{https://support.google.com/accessibility/android/answer/6122836}} follows this strategy. Specifically, the UI elements on the screen are sorted top-down and then left-right, based on their bounding box positions: \texttt{[left, top, right, bottom]}.

\subsubsection{Heuristic Methods} 
\textit{Recency} is a strong heuristics that has been constantly used in commercial products, e.g., recently used apps or recent calls. We implemented recency to predict next element to click according to the recently clicked elements on the screen. To do so, each unique screen needs to maintain a history of previous clicks. \textit{Frequency} is another commonly used heuristics in suggesting user actions. The method predicts the element to be clicked based on how often the element has been clicked on in the past. There are two design options: one based on \textit{personal frequency} that is specific to the user, the other using the \textit{global frequency} based on the aggregated frequency across users. %For the example in Table \ref{tab:click_seq_example}, $o_5$ will be predicted as the top choice and $o_3$ be the second for the personal frequency-based method. Similarly, we let this method fall back to the spatial ordering approach when there are not enough UI elements clicked in the past.

\subsubsection{Traditional Machine Learning Methods}
For this category of methods, we experiment with \textit{Logistic Regression}, \textit{SVM}, and \textit{Naive Bayesian}. These methods have been commonplaces for machine learning models before deep models became mainstream. They have been widely used in the HCI field. One challenge to apply these methods in our problem is that they cannot easily accommodate variable-length input, which is the case for both click history and the number of objects on the screen. In our experiment, we pair the previously clicked element $o_{n,c_{n}}$ with each candidate UI element on the current screen $o_{n+1,1:m}$. We featurize an element by concatenating the one-hot vector of each of these UI features: text, type, left, top, right, bottom, day-of-week, hour-of-day, and app name. We then concatenate the feature representation of both the previous and the candidate element along the one-hot encoding of the current hour and day. This forms $x_{n+1,j}$, where $1\leq{j}\leq{|s_{i}|}$, a %130,931-sized
long binary vector that is the input to, $\Phi$, either a Logistic Regression (LR), a Naive Bayesian (NB), or a SVM. The target function, $\Phi^{*}$, that we want to learn for each model is the following, where $N$ is $0$ for LR and NB and $-1$ for SVM.
\[
    \Phi^{*}(x_{n+1,j})= 
\begin{cases}
    \mbox{1},& \text{if } j=c_{i}\\
    \mbox{N},& \text{otherwise}
\end{cases}
\]

\subsubsection{Recent Deep Learning Methods}
We compared our method with three recent approaches that used deep learning methods for click prediction. Liu et al \cite{liu2015convolutional} leverage CNN in click prediction. Embedding of elements in history compose a 2D array then apply 1-D row-wise convolution, a flexible p-max pooling, and tanh as activation function. We apply a similar approach on our dataset. Lee et al. \cite{8470114} employed \textit{LSTM} \cite{10.1162/neco.1997.9.8.1735} to model mobile click sequences. A critical step in the previous approach is to tokenize each element by identifying the element using its features. Based on the previous paper \cite{8470114}, we tried our best to replicate the approach in our experiment. Each UI element is represented as a string identifier by concatenating its features such as the text content, type (as a string), and the name of the app the element belongs to. The string identifier is then treated as a token and mapped to an integer ID. With each element tokenized and represented an integer ID, it is straightforward to apply an LSTM to modeling a sequence of tokens. In addition, we compare our method with the approach introduced by Chen et al. \cite{chen2019behavior} that used a Transformer-based model for user click prediction in a shopping app. It has several critical differences with our method as we have elaborated in the Related Work section. We replicate this previous method for next click prediction in our context.

\subsection{Evaluation Metrics}
We compute multiple metrics to evaluate the prediction quality of each method. 
%For the current screen with $m$ UI objects $obj_{1:m}$, the model outputs a prediction sequence $(obj_{l_1}, obj_{l_2}, ... , obj_{l_m})$, in which $(l_1, l_2, ... , l_m)$ is a permutation of $[1, m]$. The sorting of the prediction sequence is based on the predicted possibilities of each UI objects to be clicked by the user next. We assume that in the prediction sequence $obj_{l_{1:m}}$,  the UI object user clicked on, $obj_c$, stays in position ${t}$, where $obj_{l_t}=obj_c$.

\textbf{Top-K Accuracy.} Similar to a typical model evaluation setting, we compute the top-K accuracy based on how often the target element is within the top-K predictions of a model. Particularly, we report \textit{top-1} and \textit{top-3} accuracy of each model in the experiments. For these measure, the larger the better. 

\textbf{Absolute \& Relative Ranking Position.} %We also measure how the prediction quality would potentially impact accessibility. 
Another important indicator of prediction quality is the ranking position of the target item, referred as \textit{Absolute Ranking}. As we will discuss later, for accessibility scenarios such as Android Switch access\footnote{\url{https://support.google.com/accessibility/android/answer/6122836}}, which requires a user to sequentially iterate over candidate items, the ranking position would greatly affect user experience. However, Absolute Ranking can be misleading as the difficulty to predict a target item varies based on the number of UI elements on the screen (see Figure \ref{fig:screen_distribution}). It is easier for a model to acquire a smaller Absolute Ranking on screens with few elements than on those with many elements. So we propose another measure \textit{Relative Ranking} by normalizing Absolute Ranking by the total number of elements on the screen, which is in the range of $[0, 1)$. Relative Ranking for a \textit{random} model is expected to be $0.5$. For both ranking measures, the smaller the better.

%Because each model ranks the UI elements on the screen, the quality of ranking determines the number of iterations needed from the user to locate a target element. Specifically, when using an accessibility service such as Switch Access$^{2}$, it is the number of button presses (keystrokes) needed on the attached external device to reach a target element, which we refer to as the \textit{Absolute \#Keystrokes}. It is essentially the ranking position of the target element output by the model. 

%However, Absolute \#Keystrokes can be misleading when the number of elements on a screen is small. As discussed earlier, each mobile screen has the variable number of UI elements (see Figure \ref{fig:screen_distribution}). It is easier for a model to acquire a smaller Absolute \#Keystroke on screens with fewer elements than on those with more elements. So we propose another measure \textit{Relative \#Keystrokes} by normalizing Absolute \#Keystrokes by the total number of elements on the screen, which is in the range of $[0, 1)$. Relative \#Keystrokes for a model that makes random choices is expected to be $0.5$. For both keystroke measures, the smaller the better.

\subsection{Model Configurations and Training}

We implemented all the models using Python and TensorFlow\footnote{\url{https://www.tensorflow.org}}. For Logistic Regression, SVM or Naive Bayes, we randomly sample a pair of adjacent screens from a click sequence in the training data to feed to the model. The LR model is trained by minimizing the cross entropy loss using GradientDescentOptimizer in TensorFlow. The SVM model uses SGDRegressor and the Naive Bayesian model uses BernoulliNB of the sklearn library\footnote{\url{https://scikit-learn.org/stable}}, which provides \texttt{partial\_fit()} for training on a large-scale dataset in batches. We observed that positive and negative examples are unbalanced because each screen pair only produces one positive example, since only one element is clicked, but $|s_{n+1}|-1$ negative examples. To address this issue, we gave a larger weight (5 in our experiments) to positive examples during training.

Both the CNN baseline, LSTM baseline and our Transformer-based models take variable-length click history as the input to the model. Because the length of a sequence can vary drastically, %(see Figure \ref{fig:seqence_distirbution}), 
it is not memory efficient to batch train on these sequences directly, which incurs a large number of padding. To address this issue, for the training data, we split each sequence into segments of a fixed size of 100. 
%In the Transformer Encoder for Click Sequence (see Figure \ref{fig:model}), we randomly select a sub-sequence of length $n+1$ within the chopped sequence and feed as input. Too few previous screen will fail to provide enough information for the prediction, while too long history benefits the model with little improvement in accuracy but slows down the training process or even overflows the memory. We select $n=9$ and predict the $10$th screen. 
We used a batch size of 128 and a hidden size of 128 for these models. We also tune the hyper parameters of these models such as the number of layers and the learning rate. We chose to use 2 layers for Transformer Encoder models as well as our pointer generator model (Equation \ref{eq:pointer_generator}). Adding more layers or using a larger hidden size does not significantly improve the accuracy but slows down training. All the neural models including Transformer are trained using the Adam optimizer, with a variable learning rate involving a linear warm-up followed by a exponential decay. We used a dropout ratio of 10\%. Each deep model was trained to convergence on a single Nvidia Tesla V100 GPU with 80GB memory, which took roughly 1 day to complete.

\begin{table*}[h]
  \centering\captionsetup{width=.95\linewidth}
  \begin{tabular}{l | c c c c}
    % \toprule
    % & {} & {\small \textit{Second}} & {\small \textit{Final}} \\
    Model & Top-1 Accuracy & Top-3 Accuracy & Absolute Ranking & Relative Ranking \\
    \midrule
    %Spatial & .10 & .24 & 9.04 & .48 \\

    % 4  version
    Recency & .2167 & .3573 & 7.732 & .4023 \\
    Frequency & .2314 & .3684 & 7.648 & .3970 \\
    Global Frequency & .1179 & .2633 & 8.777 & .4667 \\%[1ex]
    
    Logistic Regression & .2704 & .5327 & 4.839 & .2514 \\
    SVM & .2196 & .3925 & 7.250 & .3684 \\
    Naive Bayes & .2599 & .4905 & 4.916 & .2503 \\%[1ex]

    LSTM & .3453 & .5402 & 5.099 & .2663 \\
    CNN [Liu et al.] & .3847 & .6562 & 3.441 & .1777 \\
    Transformer [Chen et al.] & .3761 & .6463 & 3.404 & .1780 \\

    Our Model & \textbf{.4828} & \textbf{.7140} & \textbf{2.607} & \textbf{.1397} \\

    % Old version 
    % Recency & .22 & .36 & 7.73 & .40 \\
    % Frequency & .23 & .37 & 7.65 & .40 \\
    % Global Frequency & .12 & .26 & 8.78 & .47 \\
    % Logistic Regression & .27 & .53 & 4.84 & .25 \\
    % SVM & .21 & .41 & 7.03 & .36 \\
    % Naive Bayes & .25 & .48 & 5.34 & .27 \\
    % LSTM & .35 & .54 & 5.22 & .27 \\
    % % CNN & - & - & - & - \\
    % Transformer [Chen et al.] & .38 & .64 & 3.50 & .18 \\
    \midrule
    
    Our Model (all features) & .4817 & .7119 & 2.630 & .1407 \\
    Our Model (without text) & .2865 & .5370 & 4.447 & .2370 \\
    Our Model (without type) & .4777 & .7087 & 2.673 & .1428 \\
    Our Model (without position) & .4828 & .7140 & 2.607 & .1397 \\
    Our Model (without time) & .4711 & .7037 & 2.728 & .1452 \\
    Our Model (without app name) & .4780 & .7077 & 2.682 & .1431 \\
    Our Model (without in-screen attn) & .4373 & .6747 & 3.028 & .1560 \\
    \bottomrule
  \end{tabular}
  \caption{The accuracy of each model on the four metrics. For Top-1 and Top-3 accuracy, the larger the better, and for Absolute and Relative Ranking, the smaller the better. The results of the champion model is highlighted in bold.}
  \label{tab:model_result}
\end{table*}

\subsection{Test Results \& Analysis}
\subsubsection{Experimental Results}
We evaluate each model on the test dataset, by testing on every click event in each test sequence. The results are shown in Table \ref{tab:model_result}. Overall, our models outperformed all the baseline methods with a significant margin across all the metrics. Our model predicts the target element as the top choice 48\% of time and, within the top-3 predictions 71\% of time. The ranking position is improved from above 7 of heuristic methods to less than 3. 

The method using personal click frequency (Frequency) is consistently better than the one using click frequency aggregated across users (Global Frequency), which indicates that click behaviors are often personal. Recency and Frequency acquired a similar accuracy across the metrics. Machine learning-based methods generally outperformed heuristics-based ones, and deep models such as LSTM and Transformer-based approaches further outperformed traditional methods including Logistic Regression, SVM, and Naive Bayes.

Compared with the previous Transformer-based method \cite{chen2019behavior}, our model shows improvement across all the metrics. These results indicate that screen encoding enhances the learning of user behavior---Instead of only considering the clicked items, all the UI elements in the context contribute to the prediction. To further compare our model with this previous method \cite{chen2019behavior}, we split the dataset by time that was done in \cite{chen2019behavior}: for each user, the first 80\% of the sequence is used for training, the next 10\% for evaluation and the last 10\% for testing. Splitting by time yields slightly better performance for both methods than splitting by users, because a model has an opportunity to learn specific behavior of each individual user. Our model achieves 51\% and 74\% for top-1 and top-3 accuracy and 2.38 and 0.12 for absolute and relative rankings. Chen et al.'s model achieves 42\% and 68\% for top-1 and top-3 accuracy and 3.18 and 0.14 for absolute and relative rankings. Our model still substantially outperformed the previous method.

% DEPRECATED: We also experiments the Transformer model without the time encodings or package name encoding. Results shows that time encoding do play an important role in prediction. the top 1 and top 3 precision drops 0.03 and 0.02 respectively if we remove time related encodings. Surprisingly, removing package name encodings decrease the accuracy with slightly amount (<0.01). Actually this removes the requirement that a global package name list need to be known in advance. This shows that the screen content itself contains sufficient information the model need to understand the user behavior, regardless of the awareness of which app it belongs to.

% FYI, the ST V.S. STP result is at
% https://docs.google.com/spreadsheets/d/1XXNMbUw-VJpNK6nW1_9Y7ZPJJV2Kd0RJQaV1k2ZBeJY/edit#gid=555114909

%\section{Analysis \& Examples}
%\label{sec:analysis}
% - analyze experimental results
% - analyze the model behaviors

\begin{figure*}[t]
    \centering\captionsetup{width=.9\linewidth}
    \subfloat{\includegraphics[width=0.30\linewidth]{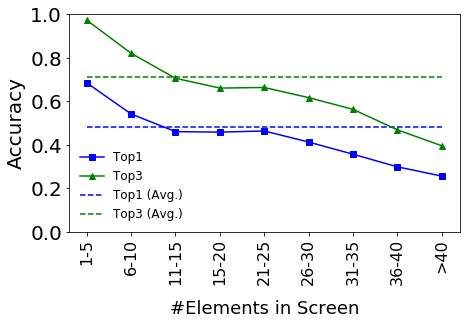}}
    \subfloat{\includegraphics[width=0.34\linewidth]{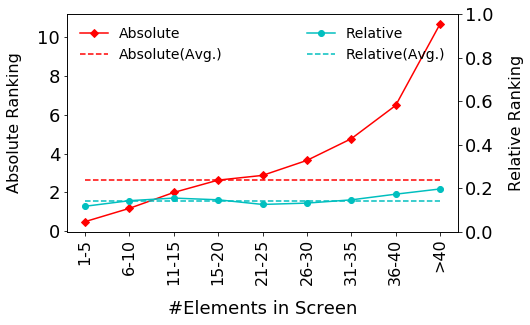}}
    \caption{Top-1 and Top-3 accuracy decrease as the number of elements on the screen increases. The Absolute Ranking increases as the number of elements on the screen increases while the Relative Ranking remains stable.}
    \label{fig:accuracy_by_screen_size}
\end{figure*}

\begin{figure*}[t]
    \centering
    \includegraphics[width=0.69\linewidth]{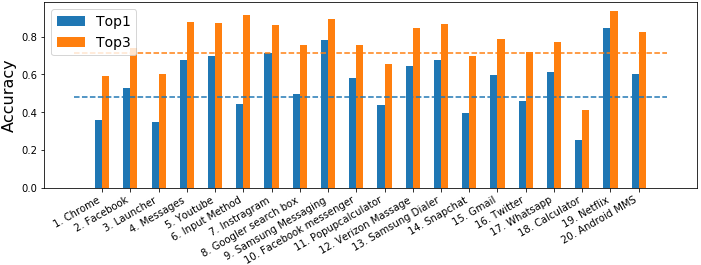}
    \caption{Top-1 and Top-3 accuracy for the top 20 popular apps in the dataset.}
    \label{fig:accuracy_by_app}
\end{figure*}

\subsubsection{Model Behavior Analysis}
% removed by zhouxin for RecSys, and extand the logigc to a full section: "Ablation Study"
% We examined two variations of our model: one using the app name $a_i$ and the other not. App name information can potentially provide more signals to the model but at the same time limit the model capability on predicting clicks for unseen apps. Based on our experiments, we found that although using the app name $a_i$ results in a small improvement on accuracy, the gain is not statistically significant for all the metrics, based on T-Test over test results from five runs of training each model variation ($p>0.05$). This result indicated our model is agnostic to app specific information and provide a general solution to model click behaviors on arbitrary UIs. 

As expected, the more elements there are on a screen the more difficult it is for a model to achieve good accuracy. As shown in Figure \ref{fig:accuracy_by_screen_size} , the top-K accuracy decreases and the ranking position is worsen as the number of elements increases on the screen. For UIs with at most 10 elements (28\% of screens), our model achieves 61\% and 90\% for top-1 and top-3 accuracy. The Absolute Ranking grows as the number of elements on the screen increases. However, the Relative Ranking remains stable regardless of the screen complexity.

To understand how the model accuracy varies across different apps, we analyzed the top-1 and top-3 accuracy for the top 20 popular apps in the dataset (see Figure \ref{fig:accuracy_by_app}). The dashed lines show the average of all apps. These 20 apps contributes 56\% of all the screens in the dataset. 13 of them show better accuracy on top-1 than the average and 16 for top-3, which indicates that model performs better on popular apps. Among these apps, the model performed the worst for the Calculator app, which is expected because the user can enter arbitrary sequences (e.g., a multi-digit number) in the Calculator that is hardly predictable.

In 26\% cross-app scenarios, i.e. $a_{n+1} \neq a_n$, the model achieves 49.6\% and 73.1\% in top-1 and top-3 accuracy, which is slightly better than the overall accuracy. This result shows that the model has robust performance in both in-app and cross-app scenarios. There's still room for improvement in prediction of in-app categories.

\begin{figure*}
\centering
\begin{minipage}{.45\textwidth}
  \centering\captionsetup{width=.95\linewidth}
  \includegraphics[width=0.44\linewidth]{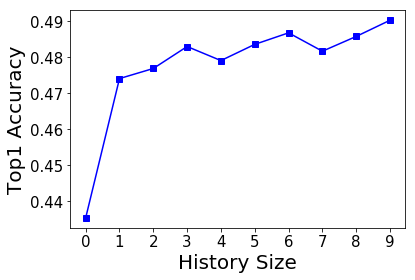}
  \includegraphics[width=0.45\linewidth]{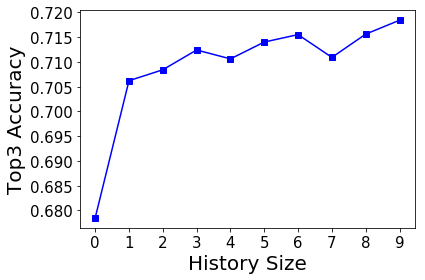}
  \newline
  \includegraphics[width=0.43\linewidth]{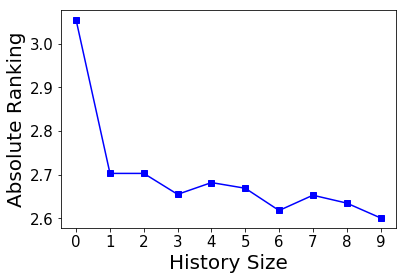}
  \includegraphics[width=0.45\linewidth]{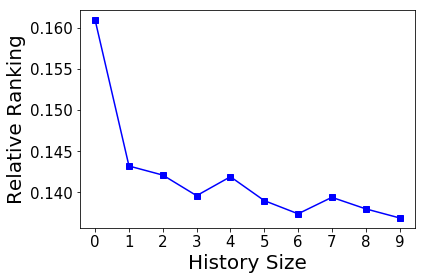}
  \caption{Performance of our Transformer model variants with different history size in model input. The model with access to the most recent screen (history\_size=1) performs 4\% better in top-1 accuracy than without history info (history\_size=0), and other metrics have similar increasement.}
  \label{fig:history_size}
\end{minipage}
\begin{minipage}{.53\textwidth}
  \centering\captionsetup{width=.95\linewidth}
  \includegraphics[width=1\linewidth]{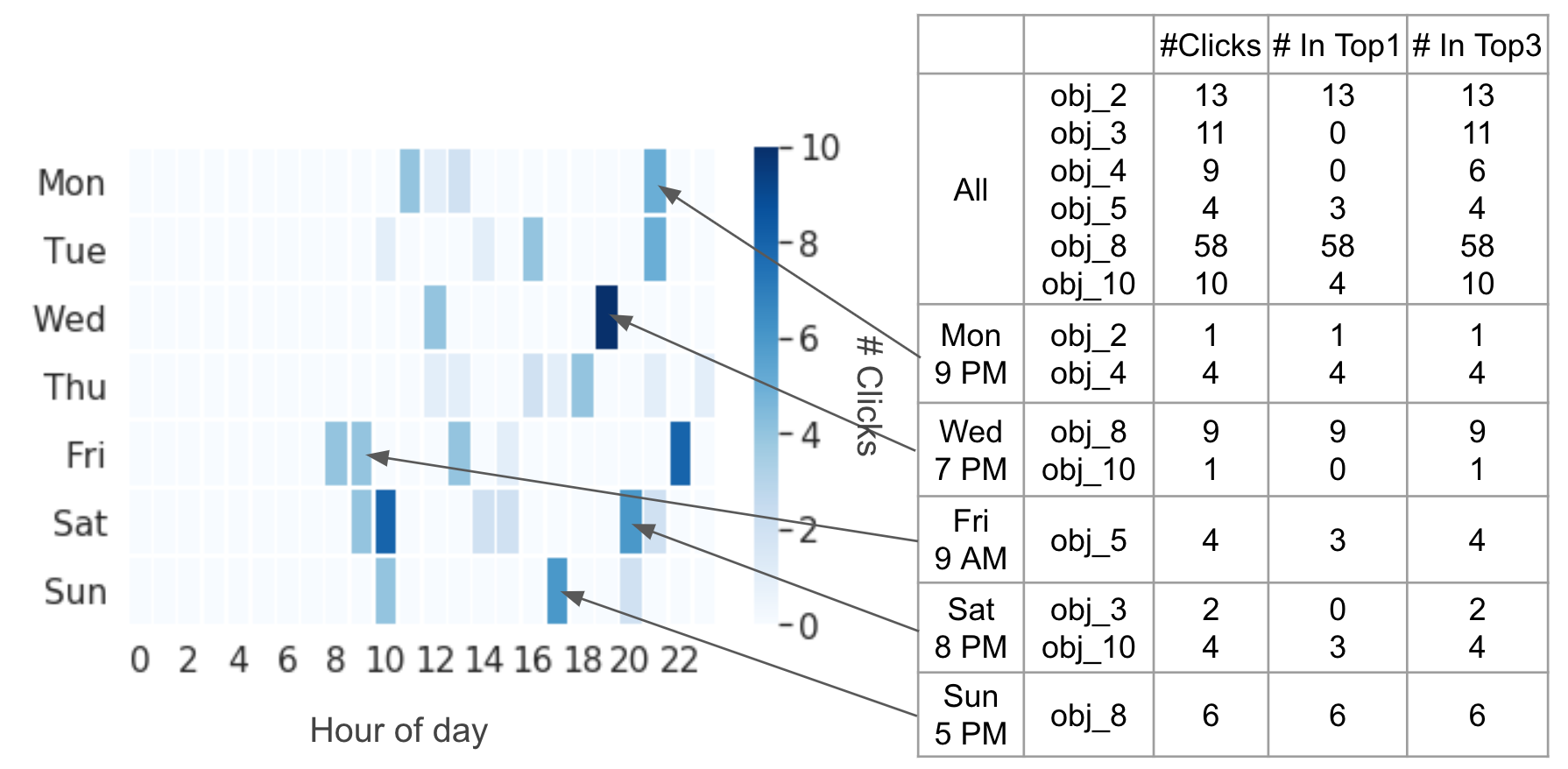}
  \caption{An analysis of click behaviors on a specific test screen that contains 12 UI elements. 6 of them were clicked by users: obj\_2, obj\_3, obj\_4, obj\_5, obj\_8, and obj\_10, which account for a total of 105 clicks on different days and hours. The heatmap on the left visualizes the distribution of these clicks across days and hours, with the color intensity corresponding the number of clicks. The table on the right elaborates on a few dense spots.}
  \label{fig:example}
\end{minipage}
\end{figure*}

To further understand the behavior of our model, we look into how our model predicts elements on specific screens. Our testing dataset contains nearly 2 millions screens, and Figure \ref{fig:example} shows the behavior of our model with respect to a specific screen that occurs 105 times. The screen has 12 UI elements, and 6 of these elements were clicked by users on different days and at different hours. These clicks are visualized by the heatmap in Figure \ref{fig:example}. As shown in the table on the right, the user clicked on different UI elements at different or the same times and our model is able to handle the variability of user behavior reasonably well. The Top-1 and Top-3 accuracy for this screen is 74\% and 97\% respectively. By breaking down the prediction by elements, we observe that obj\_2 was clicked for 13 times and our model gives perfect prediction for all of these cases. For obj\_5, our model outputs the correct prediction as the top-1 choice for 3 times out of 4 cases in total, and is able to cover all 4 cases within the top-3 predictions. 

%\begin{figure}
%\subfloat[Model performs with top1=39\%, top3=56\% for Chrome App, the most popular app in our dataset, (Test on Sun, 8PM)]{\includegraphics[width=0.2\columnwidth]{figures/example_chrome_boxes.png}} 
%\qquad
%\subfloat[Model performs with top1=59\%, top3=78\% for Gamil, (Test on Mon, 10AM)]{\includegraphics[width=0.2\columnwidth]{figures/example_gmail_boxes.png}} 
%\qquad
%\subfloat[Model performs bad for Calculator, which user clicks are hard to predict, (Test on Fri, 9PM)]{\includegraphics[width=0.2\columnwidth]{figures/example_calculator_boxes.png}} 
%\caption{Examples for model prediction on some popular apps.}
%\end{figure}

\subsection{Ablation Study}

We seek to understand how each UI feature and context affects the model performances by building variants that: 1) exclude one feature for each, 2) access different lengths of history as model input.

We experiment model variants by picking out each of these features: element text, element type, element position, time, app name. The results (at bottom part of table \ref{tab:model_result}) shows that most features contribute to the performance, in which element text makes the biggest difference. Element position helps very little in our experiment on the validation dataset, and even decreases performance in the testing dataset, as shown in the table. A suspicion is that various devices and app versions cause UI layout differences, thus providing few info for model learning. We also test a variant that uses \textit{clicked} element embedding directly without the Screen Encoder (see Figure \ref{fig:model}), the decreased performance shows that attention to other unclicked elements on screen do help in understanding the user click behavior.

The performance of the models varies by history size. The prediction is less effective without any history (as seen in Figure \ref{fig:history_size}. Access of recent screens significantly improves the performance for all metrics, while longer history provides more improvement in performance. We choose 9 as the history size for our model by the memory limit.

\section{Discussion}
% - limitation
% - open question and plan for future work

% ================= zhouxin Comment out for RecSys ================= 
% \subsection{Enabling Predictive UIs}
% ================= zhouxin Comment out for RecSys  END ================= 

%We propose a Transformer-based deep model for predicting next element to click, and compare the model with baseline models extensively based on a large-scale dataset. The findings are valuable for creating machine learning models for improving mobile usability. 
%While the focus of this paper is computational models, we recognize that to realize the benefit of our model, it is important to integrate it into an interactive system, which will further complete our understanding about how such a model would benefit end users. Particularly, one general challenge for realizing predictive user interfaces is that UI optimization, if not carefully designed, can be adversarial to usability as the dynamic change to a UI or interaction flow can hamper users for forming experiences such as spatial memory. While integrating such a model into general interaction scenarios deserves dedicated investigation, we here discuss our preliminary effort for using our model for predictive user interfaces. 

We have deployed our model to Android devices using TF-Lite\footnote{\url{https://www.tensorflow.org/lite}}, and the optimized model size is 27MB, which is within a good range for serving on a mobile device. As a proof of concept, we created a prototype feature named \textit{Next Click Overlay} that presents the UI element that is mostly likely to be clicked at the bottom of the screen. This design does not alter the layout of an existing interface, and introduces a small amount of cognitive overload for the user to glance over the predicted item. If the prediction is correct, the user can reach the next click single-handedly. For screens that the model tends to have low accuracy, e.g., Calculator, the overlay will not been shown. 

While the feature is targeted for a general mobile interaction scenario, it is particularly relevant to accessibility. Existing accessibility services such as Android Switch Access\footnote{\url{https://support.google.com/accessibility/android/answer/6122836}} allows a user with dexterity impairments to "click" on a target by iterating through the actionable elements on the UI one by one via an external device. The traversal is typically based on the spatial ordering of the elements on the screen in a top-down and left-right manner and the average number of traversal needed is 9.04 based on our dataset. With the Next Click Overlay, the number can be potentially reduced to 2.61. The estimated gain can be validated via a user study, which is beyond the scope of the paper. %Note that examining the Next Click Overlay adds a cognitive overhead when the prediction is incorrect, which needs to be investigated by a user study with the target user population. Since our focus in this paper is computational modeling, the investigation is beyond the scope of this paper.

\section{Conclusion}
We presented a novel approach for modeling mobile user click behaviors using deep learning. We analyzed a large-scale dataset of over 20 million clicks from more than 4,000 users, which involved using over 13,000 unique apps in their daily life. %Our work is motivated by accessibility scenarios where users with dexterity impairment need to traverse the elements on the UI one by one. 
Our model predicts the UI element that is likely to be clicked by the user based on the user's click history and the current context, and it provides a general solution for modeling click behaviors on arbitrary UIs. Based on a thorough experiment that compares our approach with multiple alternative methods in the literature, our model significantly outperformed all the baselines. Our experiments also show click prediction is a challenging task. We present our preliminary efforts for integrating the model into mobile interfaces, which shows promises to substantially decrease user effort needed for acquiring next element. The design of our model, our analysis and the experiments provide valuable findings for further investigating the topic.

\begin{acks}
We would like to thank Xiang Xiao, James Stout, and Ajit Narayanan from the Google Accessibility team for their help and feedback. We would also like to thank anonymous reviewers for their feedback in improving the paper.
\end{acks}

\bibliographystyle{ACM-Reference-Format}
\bibliography{click-pred}

\end{document}